\documentclass[sigconf]{acmart}
\AtBeginDocument{%
  }

\setcopyright{acmlicensed}
\copyrightyear{2025}
\acmYear{2025}
\setcopyright{cc}
\setcctype{by}
\acmConference[CIKM '25]{Proceedings of the 34th ACM International Conference on Information and Knowledge Management}{November 10--14, 2025}{Seoul, Republic of Korea}
   \acmBooktitle{Proceedings of the 34th ACM International Conference on
Information and Knowledge Management (CIKM '25), November 10--14, 2025, Seoul, Republic of Korea}\acmDOI{10.1145/3746252.3761463} \acmISBN{979-8-4007-2040-6/2025/11}
\acmISBN{978-1-4503-XXXX-X/2018/06}

\acmBooktitle{Proceedings of the 34th ACM International Conference on Information and Knowledge Management (CIKM '25), November 10--14, 2025, Seoul, South Korea}

\begin{document}

\title{MedSEBA: Synthesizing Evidence-Based Answers Grounded in Evolving Medical Literature}



\author{Juraj Vladika}
\affiliation{%
  \institution{Technical University of Munich \\ Department of Computer Science}
  \city{Garching}
  \country{Germany}}
\email{juraj.vladika@tum.de}

\author{Florian Matthes}
\affiliation{%
 \institution{Technical University of Munich \\ Department of Computer Science}
\city{Garching}
\country{Germany}}
\email{matthes@tum.de}

\renewcommand{\shortauthors}{Vladika et al.}

\begin{abstract}
In the digital age, people often turn to the Internet in search of medical advice and recommendations. With the increasing volume of online content, it has become difficult to distinguish reliable sources from misleading information. 
Similarly, millions of medical studies are published every year, making it challenging for researchers to keep track of the latest scientific findings.
These evolving studies can reach differing conclusions, which is not reflected in traditional search tools.
To address these challenges, we introduce MedSEBA, an interactive AI-powered system for synthesizing evidence-based answers to medical questions. It utilizes the power of Large Language Models to generate coherent and expressive answers, but grounds them in trustworthy medical studies dynamically retrieved from the research database PubMed. The answers consist of key points and arguments, which can be traced back to respective studies. Notably, the platform also provides an overview of the extent to which the most relevant studies support or refute the given medical claim, and a visualization of how the research consensus evolved through time. Our user study revealed that medical experts and lay users find the system usable and helpful, and the provided answers trustworthy and informative. This makes the system well-suited for both everyday health questions and advanced research insights.

\end{abstract}




\maketitle

\section{Introduction}
In an era characterized by quick access to information, the Internet has become the primary resource for individuals seeking answers to health-related questions \cite{eurostat2020, thapa2021influence, luo2022effect}. However, this ease of access comes with a risk: differentiating between scientifically validated health advice and unsubstantiated claims poses a challenge for the general public \cite{swire2020public, teplinsky2022online}, which can potentially lead to incorrect self-diagnosis and harmful health decisions \cite{pennycook2020fighting, jabbour2023social}. The preferred source of medical guidance is rooted in credible, peer-reviewed scientific studies, but navigating the increasingly large and dynamic landscape of medical research is a practical challenge \cite{gonzalez2024landscape}. 

Even medical researchers can struggle to keep up with the latest discoveries and require tools that can easily give them an overview of how the research is developing.
Additionally, it is not uncommon that certain studies yield conflicting results \cite{carpenter2016conflicting, sylvester2017conflict}. 
Traditional search tools are ill-equipped to systematically synthesize these divergent findings, making it difficult to grasp the overall consensus on a given medical topic. Most existing tools for searching academic databases, like Google Scholar, provide an overview of relevant publications, without assessing their \textit{stance} towards the question, i.e., whether they support or refute the given hypothesis \cite{hardalov2022survey}.

To address these challenges, we introduce \textbf{MedSEBA}, a system designed to provide a  \textit{synthesis of evidence-based answers} to complex medical questions,
using Natural Language Processing (NLP) techniques and the generation capabilities of Large Language Models (LLMs) \cite{singhal2025toward}. An essential feature of our system is the reliability of its outputs, which is ensured by grounding every piece of information in the provided answers to medical studies retrieved from the popular medical academic database PubMed. Unlike many other search tools, it grounds the answers solely in medical papers and also provides direct citations to relevant sentences in sources.

The system presents users with a structured answer that includes key arguments, a consensus assessment indicating the degree to which current research supports or refutes a given claim, and charts with a temporal overview of how the research on this question has evolved. These visualizations make the system attractive for medical researchers who want to get a quick overview of what the current research says about their prospective research hypotheses.

This paper details the architecture and workflow of the MedSEBA prototype and provides results of a small-scale user study, revealing the general satisfaction with the system usability and answer quality, while leaving room for future improvements. A demonstration video is uploaded to YouTube,\footnote{\textbf{Video:} \url{https://youtu.be/32GZsNaychA}} while the system code can be found in a public GitHub repository.\footnote{\textbf{Code}: \url{https://github.com/jvladika/MedSEBA}}

\begin{figure*}
    \centering
    \includegraphics[width=0.66\linewidth]{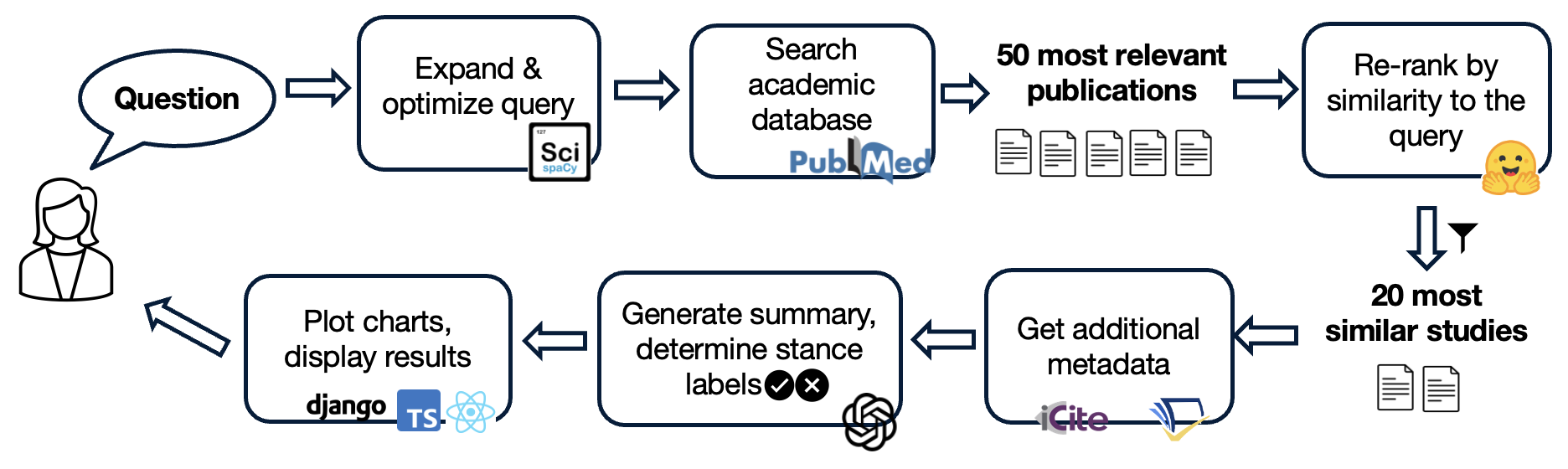}
    \caption{Workflow of the main functionality of the system -- answering and analyzing user's medical question.}
    \label{fig:workflow}
\end{figure*}

\section{Related Work and Systems}
The popular platforms for indexing general academic research include Google Scholar \cite{harzing2008google} and Semantic Scholar \cite{scholar2024semantic}. Some others focus on specific fields of research, such as NLP-KG for NLP research \cite{schopf2024nlp}, EvidenceMiner for life sciences \cite{wang2020evidenceminer}, and PubTator or LitSense for searching PubMed \cite{wei2024pubtator, yeganova2025litsense}. These databases index a vast amount of studies and can be used to quickly find literature related to the given input query. Still, they do not provide insights into the stance (support) of the studies toward the query.

Our use case is related to automated fact-checking, which aims to determine the veracity of a given claim based on credible evidence \cite{guo2022survey, vladika-matthes-2023-scientific}. There have been systems developed for real-time fact verification, like ClaimBuster \cite{hassan2017claimbuster} and CoVerifi \cite{kolluri2021coverifi}, but they usually only provide the final label (e.g., \textit{refuted}) without giving an overview of different sources and arguments. This is reasonable for clear-cut facts such as "\textit{Berlin is the capital of Germany}". On the other hand, the nature of many medical questions posed to our system, like "\textit{Does vitamin C alleviate colds?}", is that they are nuanced and medical research might have reached different conclusions over the years, or the clinical recommendations differ depending on patients' characteristics such as age or genetics.

Another similar use case is generative search, which provides long-form answers and summaries to given input prompts based on retrieved sources and references \cite{colton2021generative}. It is offered by popular services like Perplexity.ai \cite{noauthor_perplexity_nodate}, and as of recently by web versions of popular LLMs like ChatGPT \cite{roumeliotis2023chatgpt} and Gemini \cite{team2024gemini}. Still, the referenced sources are often only linked to, without highlighting which parts actually discuss the input question, as well as without providing the stance labels and clear key arguments.

\section{Architecture}
We use a standard frontend-backend system with a model-view-controller architecture and three layers. The frontend of the web application uses React and TypeScript frameworks. The backend utilizes the Python-based Django framework. The vector database used for storing cached embeddings is Weaviate, while user data and search data are stored in the relational database PostgreSQL.

\subsection{Main Workflow}
The core functionality of our knowledge assistant is to provide evidence-based answers from medical studies to posed medical questions. The system operates through a multi-stage pipeline, ensuring both broad retrieval and precise information extraction and synthesis. This is visualized in Figure \ref{fig:workflow}. 

\textbf{Document Retrieval.} After the user inputs a question, the first step is transforming it into a query optimized for PubMed's search engine. We use SciSpacy \cite{neumann2019scispacy}, a library trained on biomedical text supporting advanced text processing, to perform named entity recognition (NER), detect relevant medical concepts, and propose synonyms and related entities. This is used to expand the search query with multiple options.
For example, a query can be transformed into a Boolean search expression: \textit{predictors AND poor AND surgical outcomes AND elderly cardiac surgery AND (patients OR outpatients OR visitors to patients) AND ("journal article"[Publication Type] OR "review"[Publication Type])}. This optimized query incorporates key terms, their synonyms and related concepts, and filters for specific publication types, enhancing the specificity of the subsequent search.
After the query is finalized, we first retrieve 50 research papers deemed most relevant by PubMed's internal relevance ranking based on the query. These 50 papers serve as the candidate pool for the subsequent, more granular semantic filtering.

In the next step, the 50 papers are narrowed down to the 20 most relevant ones based on semantic similarity. The query and 50 candidate documents are embedded into a vector representation with a sentence transformer model, and the 20 most similar documents are selected based on cosine similarity to the query. The model used is BMRetriever \cite{xu-etal-2024-bmretriever}, which is optimized for biomedical text. We chose 20 documents based on experiments, where we found 10 to lack coverage, while 30 had too much noise \cite{vladika-matthes-2025-influence}.

\begin{table}[h!]
	\caption{The system prompt used for summary generation}
	\centering
    \scriptsize
	\begin{tabular}{|p{0.46\textwidth}|}
		\hline
		  Format:
    [Answer to the question based on the documents in one sentence]
    
    [Grouping of documents into clear categories with a heading; one sentence summary for each category of grouped documents with references (e.g. [1])]
     
     Example output for how it should look:
     Sitting for prolonged periods is generally considered detrimental to health, particularly when combined with low levels of physical activity, as it is associated with increased risks of chronic diseases and mortality.

\#\#\# Health Risks of Sitting

- **Sitting and Mortality**: Studies indicate that prolonged sitting is associated with increased all-cause and cardiovascular disease mortality, especially for those who are least active ([1], [4]).

- **Chronic Diseases**: Excessive sitting is linked to a higher risk of various chronic diseases, including type 2 diabetes and cardiovascular disease, reinforcing its status as an independent risk factor ([4], [7]).
		\\ \hline
	\end{tabular}
	\label{tab:prompt}
\end{table}

The final stage of the pipeline focuses on enriching the selected 20 papers and synthesizing their content into an evidence-based answer. For each of the 20 most relevant papers, external APIs are called to retrieve information that the PubMed API lacks, like the number of citations (iCite \cite{icite_hutchins_santangelo_2019}) and venue information (Semantic Scholar API \cite{kinney2023semantic}).
We are interested in showing users metadata relevant to the trustworthiness of the studies. Metadata like year of publication, the publishing venue (journal or conference), as well as the number of citations, are general indicators of the popularity and relevance of this study. Recent work has found that studies with more citations and with a more recent publication date lead to more reliable predictions of the veracity of medical claims \cite{vladika2024improving}. 

\textbf{Answer Synthesis.} Once the final set of 20 studies is chosen, a synthesized summary is generated by an LLM (GPT-4o) using a detailed prompt. This prompt instructs the model to structure the key arguments into a summary, and to support its statements by referring to exact retrieved studies (see Table \ref{tab:prompt}). The abstracts of all 20 studies are numbered and passed as input to the model.

\begin{figure}[h]
    \centering
    \includegraphics[width=0.98\linewidth]{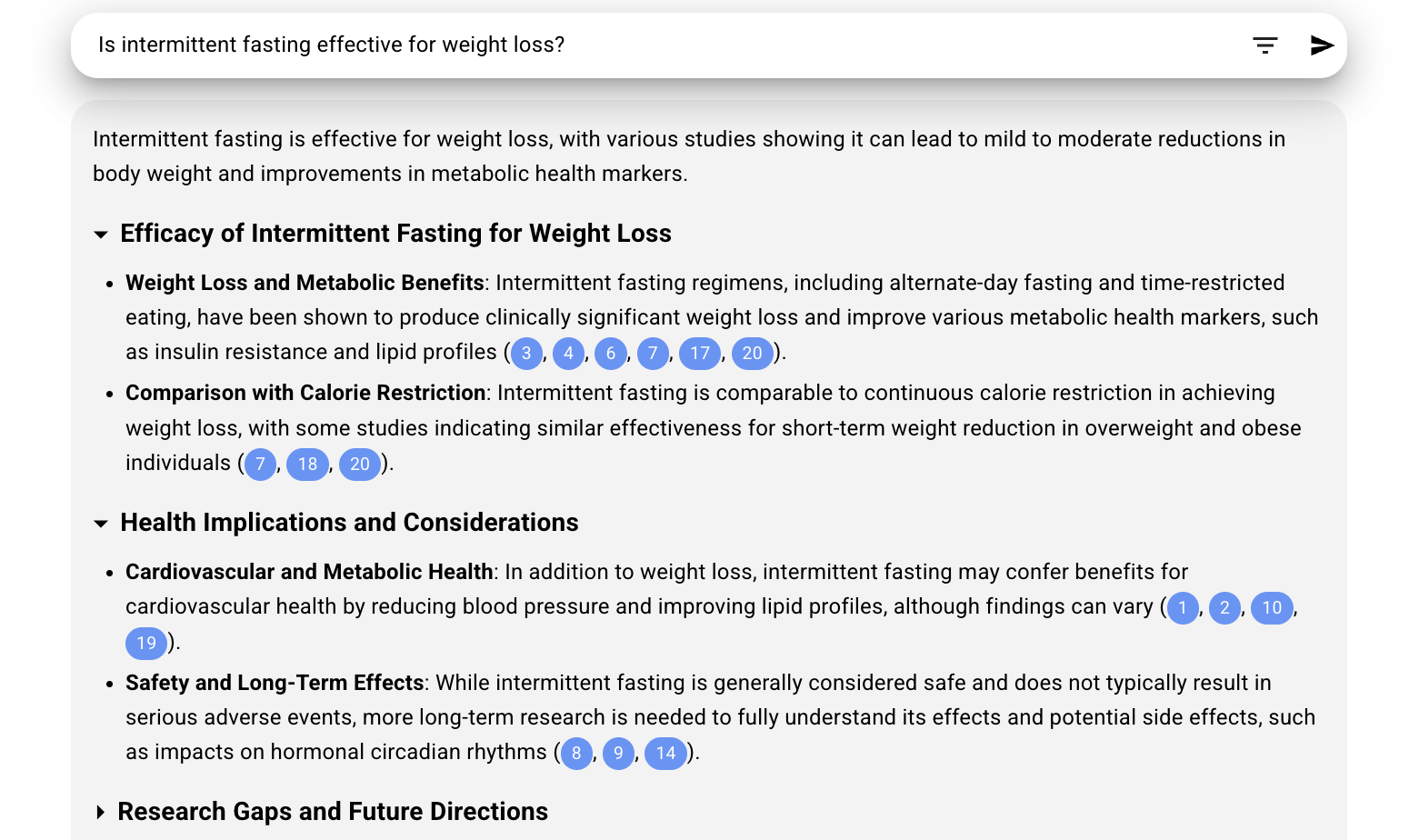}
    \caption{Generated answer to the medical question incorporates retrieved papers and outlines key points}
    \label{fig:ss_summary}
\end{figure}

\begin{figure}[h]
    \centering
    \includegraphics[width=0.86\linewidth]{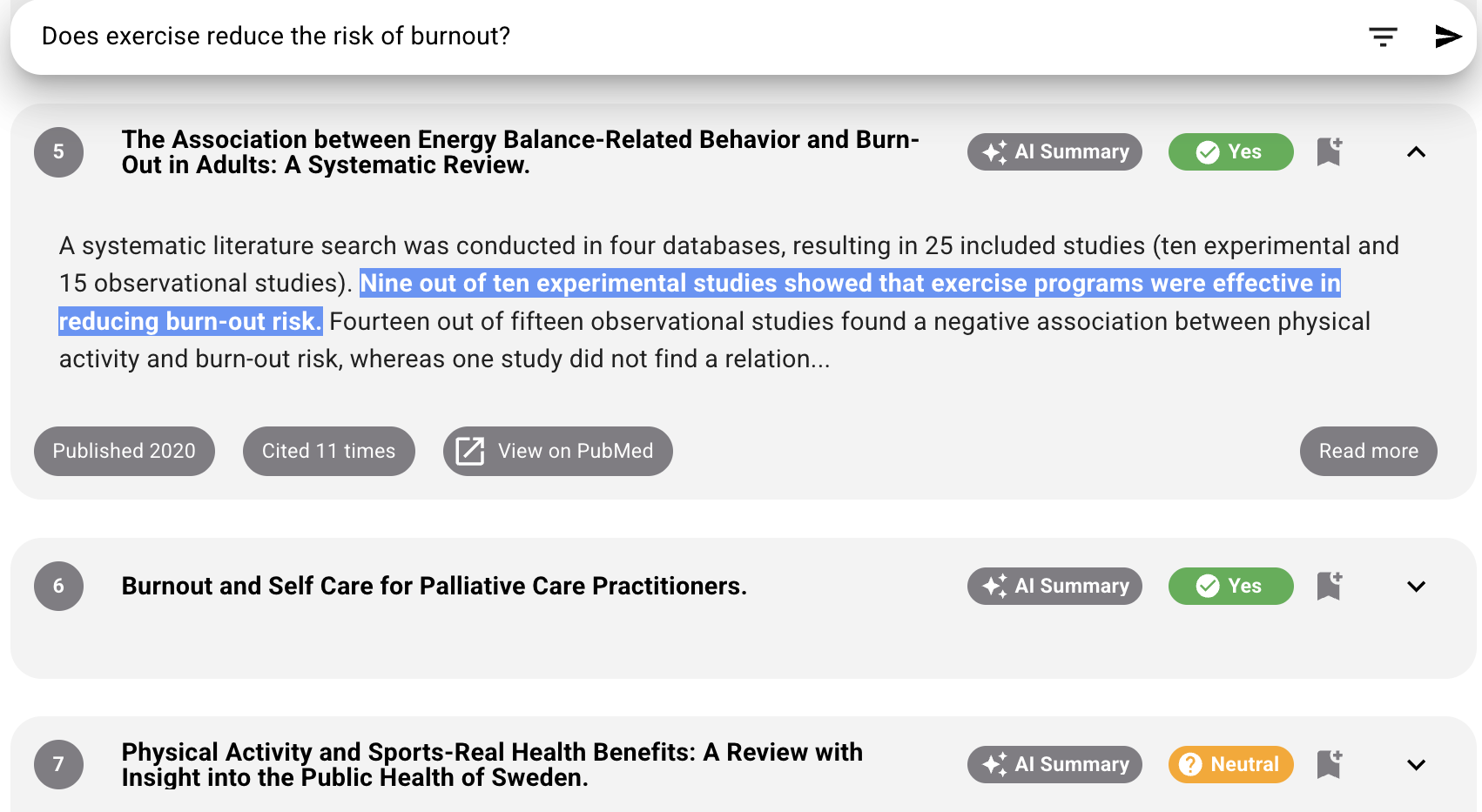}
    \caption{Retrieved papers listed below the answer, with metadata, stance, and the most relevant sentence}
    \label{fig:ss_papers}
\end{figure}

The answer is displayed with clickable references (Figure \ref{fig:ss_summary}) and the 20 studies are listed below the summary in respective boxes (Figure \ref{fig:ss_papers}). Each box includes name and abstract, the relevant metadata (year, citations, venue), the stance towards the query, a short AI-generated summary. For determining the stance, GPT-4o (2024-11-20) is prompted to assess whether the paper's findings support or refute the assertions of the question, or have no relation. While we also experimented with encoder-only models fine-tuned for Natural Language Inference (NLI), such as DeBERTa-v3 \cite{he2023debertav3improvingdebertausing}, it predicted the neutral ("not enough information") class too often, due to the label shift from its training data on this applied scenario. 
Therefore, we found GPT-4o to be a more reliable stance predictor.


Finally, the sentence embedding model BMRetriever is again employed to identify the single most relevant sentence within each paper that directly addresses the user's query. This direct evidence extraction allows for improved verifiability and precise contextualization of the broader summary generated by the LLM \cite{liu-etal-2023-evaluating}.

\textbf{Data Source.} PubMed was shown to be a reliable knowledge source for verifying scientific claims and answering medical questions \cite{goodman2023accuracy, vladika-matthes-2024-comparing}. We initially experimented with using a local PubMed dump, by storing the embeddings of all documents in a vector database and retrieving them during search. We later switched to completely using the public PubMed API to retrieve the documents. The first reason is the reduced need for hardware and storage resources, considering the vast size and calculation time of cosine similarities to all embeddings for each input query. The second reason is the timeliness of information -- a local dump gets outdated and needs to be regularly updated, while accessing PubMed through API will always provide the latest state of research.

PubMed indexes more than 38 million publications. For each publication, an abstract is available. In the biomedical domain, abstracts are often enough for getting an insight into the study's main findings \cite{wadden-etal-2020-fact}, especially because abstracts are longer and usually end with the discussion of main implications. 
For around 5 million documents, the full text is available -- we dynamically fetch these and show as an integrated PDF to users within our document view.

\subsection{Document Page}
A user can click on one of the 20 documents used for the answer generation and inspect it in more detail. 
The document page shows metadata retrieved in the previous step and the generated overarching topics (tags) of the paper. For those studies that are open-access, the PDF of the full study is retrieved from PubMed Central and shown to users in a built-in PDF viewer (based on \textit{React-pdf}). Users can take and save notes in the PDF. Currently, highlighting of specific parts is not supported and is left for future upgrades.

\begin{figure}[h]
    \centering
    \includegraphics[width=0.8\linewidth]{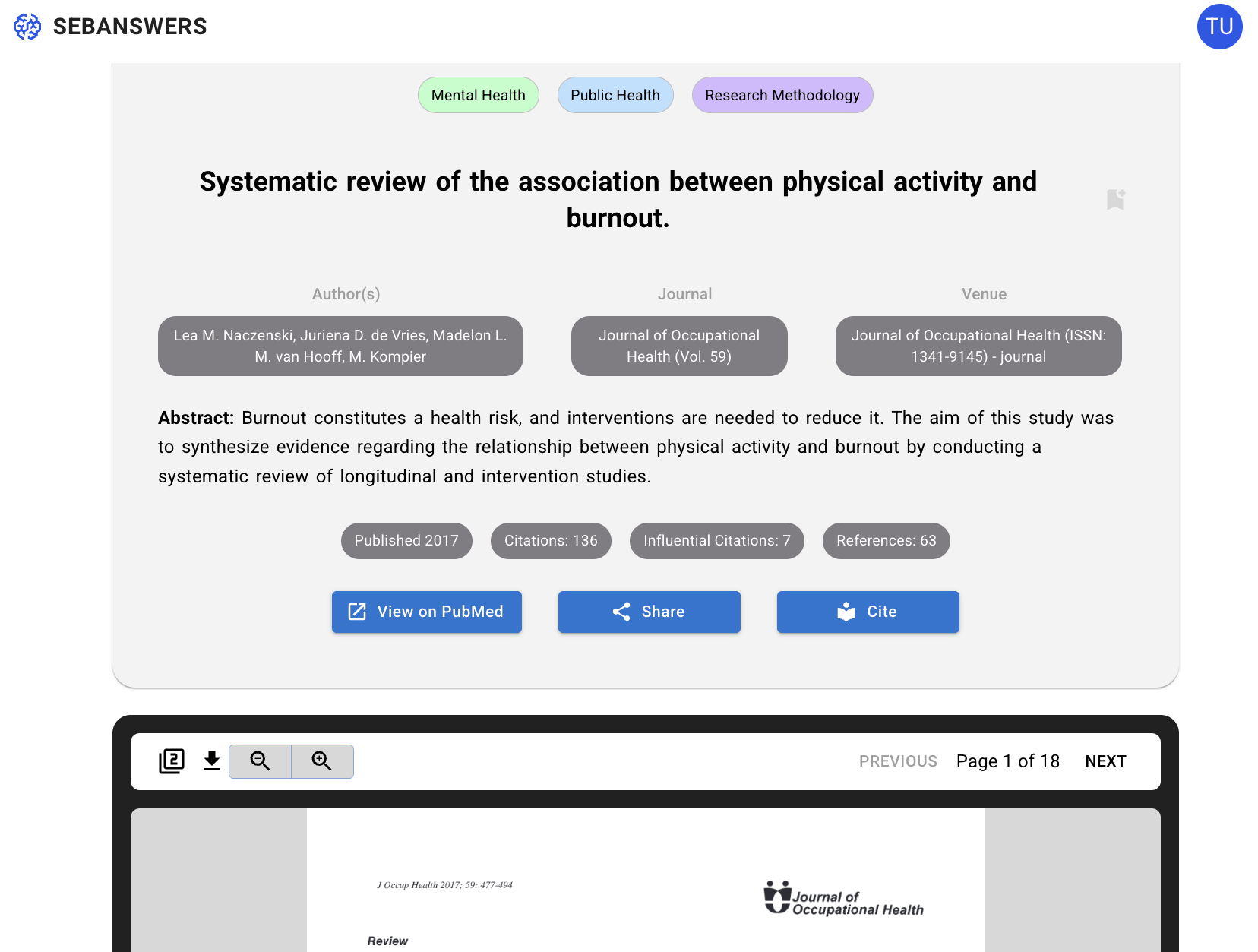}
    \caption{Document page shows more data about the paper}
    \label{fig:ss_document}
\end{figure}

\subsection{Visual Plots}
\begin{figure}[h]
    \centering
    \includegraphics[width=0.7\linewidth]{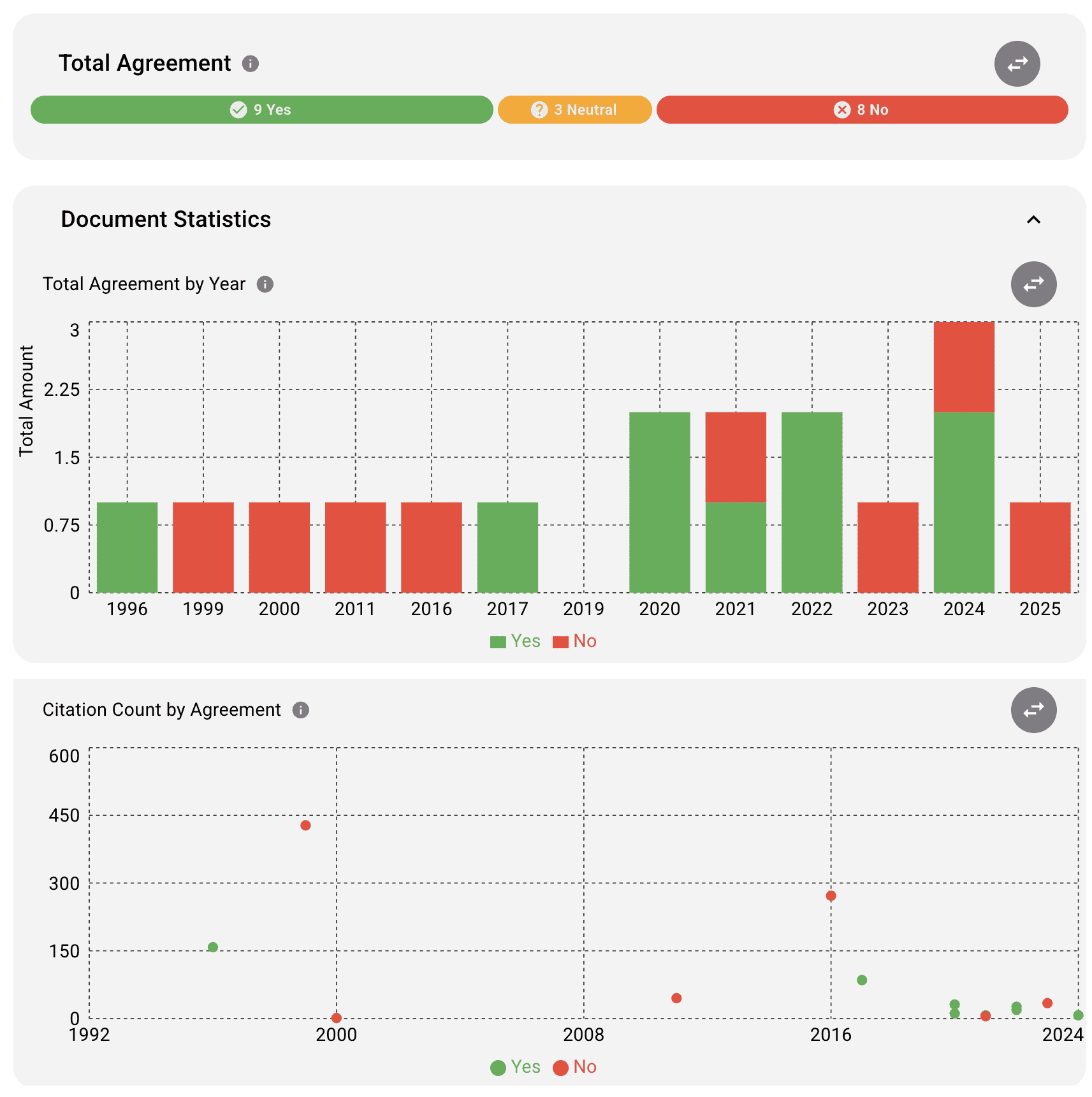}
    \caption{Visual charts help contextualize the state of research}
    \label{fig:ss_charts}
\end{figure}

There are three charts that help visualize the main search results (Figure \ref{fig:ss_charts}). The first one is a stacked bar chart that shows the distribution of stance labels (supported, refuted, neutral) among the 20 selected publications. This supports two views, either showing the final label distribution, or the distribution of the strengths of labels for each publication -- so if a study 70\% supports and 30\% refutes the query, the first view would just count the \textit{supported} label, while the second view would take the value of both into account.

The second and third charts are time series charts that show how has the research on the given hypothesis has developed through time. The second chart shows stacked bar charts with support/refute labels for each featured year. The third chart plots the citation count of each study, with the axes showing year and citation number, respectively, while the color shows the dominant stance label. 

These three charts are useful for lay users and researchers alike. In particular, this gives scientists a quick visual overview of how a certain research question has developed throughout time. For the example shown in Figure \ref{fig:ss_charts}, we see that while the initial research on this topic started with negative findings, more recent studies have found a positive relation between the observed phenomena. 

\subsection{Additional User Features} 
Our system also has features that enhance the overall user experience and usability of the system. Users can create a user account, which enables them to store their search history, group their search results into folders, and have an overview of their most commonly sought topics. Still, given that users could ask sensitive and confidential medical questions, we also offer the system to be used completely anonymously and an option to delete search history.

\section{User Study}
We evaluated the usability and reliability of the system with a small-scale user study. The study featured ten participants: two medical experts (finished medical education) and eight lay users (pursuing a graduate degree in computer science). All users are in-house colleagues; therefore, no monetary compensation was made. Instructions were provided beforehand, as well as a short description of the system. We asked the users to choose at least three questions covering different medical domains and health issues, and explore the application. Afterward, we asked them to fill out System Usability Scale (SUS) \cite{lewis2018system}, a common set of ten questions on general usability and accessibility of a software system. 

To evaluate SUS, we normalize the average score to a 0--100 range. A commonly used benchmark score of satisfying performance is 68 \cite{sauro2011practical}, which was also shown to be valid for digital health apps \cite{info:doi/10.2196/37290}. Our app obtained an average of 81.7, which is significantly better than the benchmark score of 68 (t-test, p<0.01). 

Additionally, we asked the users to evaluate six statements on the reliability of the provided information, rated on a five-point Likert scale from "strongly disagree" to "strongly agree".
Three statements focused on accuracy and three on the quality of the summary. The statements were, with percentage of positive answers ("agree" or "strongly agree") in parentheses: "\textit{I find the retrieved studies relevant for the query}" (90\%), "\textit{I find the predicted stance labels sensible}" (80\%), "\textit{I find the highlighted sentences related to the query}" (70\%); "\textit{I find the summary to be understandable}" (80\%), "\textit{I find the summary to be informative}" (90\%), "\textit{I find the study to be complete with respect to retrieved studies}" (70\%).  The overall percentages of positive answers were found to be significant (binomial test, p<0.05). 

This shows that users were overall satisfied and rated the system highly, but the weakest points of the system, and the biggest research and development challenges, are completeness of the summary (it happens due to the length of the summary that some studies are left out), and choosing the most relevant sentence in the study abstract. These will be the focus of future upgrades.

\section{Conclusion}
In this system demonstration paper, we introduce MedSEBA, an interactive system dedicated to providing synthesized, evidence-based answers to user questions related to medicine and health. The system provides detailed answers with key arguments based on medical research papers, tracing back to original sources, stances of studies towards the query, and a visual representation of the current research consensus and how it has evolved through time for the given hypothesis. In addition, the system supports full document view and structuring the search history. Our user evaluation shows a high level of system usability, quality, and answer trustworthiness, making the system well-suited for lay users seeking reliable health advice, as well as scientists exploring the latest research for their inquiries. Future work will expand the system by improving the answer completeness and source highlighting, as well as supporting locally deployed LLMs (that we omit due to resource constraints), which would be an important factor for privacy-sensitive and confidential usage of this system in clinical environments.

\section*{Acknowledgements}
This research has been supported by the German Federal Ministry of Education and Research (BMBF) grant 01IS17049 Software Campus 2.0 (TU München). 
We would like to thank Dr. Bob Schijvenaars for insightful discussions and helpful suggestions to this project. 

We also thank the students of TUM who were essential in developing the web application: Alexander Blatzheim, Stefania Mocan, Yunus Celik, Jingyu Wang (2024); and Sofiia Danylchenko, Florian Seitz, Nikolas Anton Lethaus, Henning Paul Ilbertz (2025). 

\section*{GenAI Usage Disclosure}
No Generative AI tools were used for writing this research paper. As described in the \textit{Architecture} section, Large Language Models are part of the AI pipeline used in the shown demo system.

\bibliographystyle{ACM-Reference-Format}
\bibliography{acmart}

\end{document}